\documentclass[conference]{IEEEtran}
\IEEEoverridecommandlockouts
\usepackage{kotex} 

\usepackage{cite}
\usepackage{amsmath,amssymb,amsfonts}
\usepackage{comment}
\usepackage{graphicx}
\usepackage{mathtools}
\usepackage{textcomp}
\usepackage{xcolor}
\usepackage{setspace}
\usepackage{tabularray}
\usepackage{pifont}
\usepackage{cite}
\usepackage{amsmath,amssymb,amsfonts}
\usepackage{algpseudocode}
\usepackage{graphicx}
\usepackage[final]{microtype}
\usepackage[italic]{mathastext}
\usepackage{libertine}
\usepackage[T1]{fontenc}
\usepackage{textcomp}
\usepackage[varqu,varl]{zi4}
\usepackage[all]{nowidow}
\usepackage{setspace}
\usepackage{colortbl}
\usepackage{url}
\usepackage[hyphenbreaks]{breakurl}

\usepackage{textcomp}
\usepackage{xcolor}
\def\BibTeX{{\rm B\kern-.05em{\sc i\kern-.025em b}\kern-.08em
    T\kern-.1667em\lower.7ex\hbox{E}\kern-.125emX}}

\usepackage[utf8]{inputenc}
\usepackage{kotex}
\usepackage{color}
\usepackage[square,numbers,comma,sort]{natbib}
\usepackage[linesnumbered,ruled,vlined]{algorithm2e}
\usepackage[normalem]{ulem}
    
\usepackage{makecell}
\usepackage{dblfloatfix} 
\usepackage{lipsum}
\usepackage{booktabs}
\usepackage{multirow}
\usepackage{amssymb}
\usepackage{graphicx}
\usepackage{siunitx}
\usepackage{verbatim}

\usepackage{ulem}
\usepackage{url}

\definecolor{mygreen}{RGB}{0,128,0}

\newcommand{\cb}{\textcolor{blue}}

\newcommand{\squishlist}{
\begin{list}{$\bullet$}
	{ \setlength{\itemsep}{0pt}      \setlength{\parsep}{-0pt}
		\setlength{\topsep}{4pt}       \setlength{\partopsep}{0pt}
		\setlength{\listparindent}{-2pt}
		\setlength{\itemindent}{-5pt}
		\setlength{\leftmargin}{1em} \setlength{\labelwidth}{0em}
		\setlength{\labelsep}{0.5em} } }

\newcommand{\squishend}{
\end{list}}

\newcommand{\tbdfixed}{\texttt{InferSave}}

\begin{document}

\title{
Cost-Efficient LLM Serving in the Cloud: VM Selection with KV Cache Offloading
}
\author{\IEEEauthorblockN{
Kihyun Kim$^{1}$, Jinwoo Kim$^{1}$, Hyunsun Chung$^{1}$, Myung-Hoon Cha$^2$, 
Hong-Yeon Kim$^2$, Youngjae Kim$^{1,\dagger{}}$
\thanks{$^{\dagger}$Y. Kim is the corresponding author.}
}
\IEEEauthorblockA{$^1$Dept. of Computer Science and Engineering, Sogang University, Seoul, Republic of Korea\\
$^2$ETRI, Daejeon, Republic of Korea} 
}
\maketitle


\begin{abstract}
LLM inference is essential for applications like text summarization, translation, and data analysis, but the high cost of GPU instances from Cloud Service Providers (CSPs) like AWS is a major burden.
This paper proposes \tbdfixed{}, a cost-efficient VM selection framework for cloud-based LLM inference. \tbdfixed{} optimizes KV cache offloading based on Service Level Objectives (SLOs) and workload characteristics, estimating GPU memory needs, and recommending cost-effective VM instances. Additionally, the Compute Time Calibration Function (CTCF) improves instance selection accuracy by adjusting for discrepancies between theoretical and actual GPU performance.
Experiments on AWS GPU instances show that selecting lower-cost instances without KV cache offloading improves cost efficiency by up to 73.7\% for online workloads, while KV cache offloading saves up to 20.19\% for offline workloads.

\end{abstract}

\section{Introduction}
\label{sec:intro}

Large Language Models (LLMs) have become a core technology in modern Natural Language Processing (NLP), demonstrating outstanding performance in various applications such as text summarization, machine translation, and conversational AI~\cite{vaswani2017attention}. LLMs built on Transformer-based architectures, such as GPT~\cite{radford2018improving} and LLaMA~\cite{touvron2023llama}, leverage multi-layer self-attention mechanisms and large-scale pretraining to achieve near-human-level language understanding and generation capabilities. Thanks to their superior performance, LLMs are widely used across industries, providing high accuracy and natural responses in a wide range of tasks, including text summarization, question answering, and document analysis.

However, to efficiently design an LLM inference system, it is essential to consider task-specific Service Level Objectives (SLOs). For instance, in online inference tasks, such as real-time conversational services or question answering, latency must be minimized to ensure a seamless user experience. Reducing inference latency is a key challenge in these scenarios.

On the other hand, in batch processing tasks~\cite{anthropic2023message, openai2024batch} such as text summarization for large datasets, log analysis, and document clustering, latency requirements are generally less strict. Instead, maximizing throughput is critical, as these tasks involve processing large volumes of input data at once. In such batch processing environments, handling large batches can easily lead to GPU memory shortages. Due to the auto-regressive nature of LLM inference, the Key-Value (KV) cache, which stores past token information, continuously grows. As a result, GPU memory usage increases sharply with sequence length and batch size.

A common technique to mitigate this issue is KV Cache Offloading, which offloads KV cache data exceeding GPU memory limits to CPU memory or disk. This enables large-batch processing without running out of memory~\cite{flexgen,layerkv,instinferinstorageattentionoffloading,Deepspeed}. However, if the additional latency introduced by offloading is not properly managed, throughput can significantly degrade, potentially failing to meet the required SLOs.

\textbf{Cost Efficiency of LLM Inference in Cloud Environments:} 
Major cloud service providers such as AWS, GCP, and Azure offer a variety of GPU instance options with different performance levels and cost structures, providing flexibility in resource utilization~\cite{google_cloud_comparison}. However, selecting a cost-efficient GPU instance in a cloud environment is a complex task that is difficult for users to perform manually. The challenge arises because GPU instances vary significantly in price and performance (Refer to Table~\ref{tbl:awstypes}), and workload characteristics require flexible KV cache offloading strategies, making optimal selection difficult.

Given this complexity, an optimized approach must integrate the following two key factors:
\begin{itemize}
\item GPU instance selection based on task characteristics
\item Efficient KV Cache Offloading strategy
\end{itemize}

Balancing throughput targets and cost efficiency by combining these two factors remains a critical challenge that needs to be addressed.

\textbf{Limitations of Existing Research:}
Previous studies on cost efficiency in cloud environments~\cite{Proteus,deepvm,clustering-analysis2023,andrzejak2010decision,kokkinos2013cost} have focused primarily on image processing or general machine learning workloads. As a result, they do not capture the unique characteristics of large-scale LLM inference. Moreover, recent research on cost-efficient LLM inference has largely concentrated on real-time inference scenarios~\cite{melange, aladdin,thunderserve, splitwise,wang2024revisiting}, neglecting large-scale data processing environments where KV cache offloading could be leveraged effectively. Furthermore, these studies do not comprehensively analyze cost efficiency in relation to Service Level Objectives (SLOs).

To address these challenges, this paper proposes \tbdfixed{}, a software framework that automatically selects the optimal VM instance by considering both cost and performance based on SLOs.

The \tbdfixed{} framework operates as follows based on user input:
First, It calculates the required GPU memory based on the specified SLO and workload size, analyzing the feasibility of KV cache offloading to determine a set of candidate instances. Next, using pre-collected performance data, it performs a modeling step to predict the performance and cost of each instance. Finally, it evaluates these predictions to recommend the most cost-efficient instance that meets the user's SLO constraints.Through this process, the InferSave framework becomes the first solver system that automatically recommends the most economical VM instance for LLM serving in cloud environments. By integrating KV cache offloading and GPU instance characteristics, it ensures SLO compliance while optimizing costs.

The \tbdfixed{} framework analyzes GPU instance performance based on user input and comprehensively considers the feasibility of KV cache offloading to automatically recommend the optimal VM instance for LLM inference in cloud environments. By leveraging \tbdfixed{}, users can easily find the most cost-effective VM instance that meets their specified SLO while minimizing operational expenses.

Experimental results show that applying \tbdfixed{} achieves significant cost savings compared to traditional maximum-performance-based policies, with reductions of up to 73.7\% for online workloads and 20.19\% for offline workloads. In addition, it is designed to be flexible across various AWS instances and cloud environments, providing a practical and efficient approach to operating LLM inference services.

\section{Background and Motivation}
\label{sec:background}

\subsection{LLM Architecutre and Inference}

Large-scale language models (LLMs), such as OpenAI’s GPT~\cite{radford2018improving} and Meta’s LLaMA~\cite{touvron2023llama}, are built on the Transformer~\cite{vaswani2017attention} architecture. These models consist of a multi-layer structure incorporating Self-Attention mechanisms and Feed-Forward Networks, enabling their broad applicability across various natural language processing (NLP) tasks.

The LLM inference process is divided into two stages: Prefill and Decode. In the Prefill stage, the input prompt is processed in parallel to generate the initial output tokens. During this process, Query, Key, and Value vectors are computed for each token in the input prompt, capturing contextual information through token-wise interactions. Simultaneously, the computed Key and Value tensors are stored in the GPU memory as a Key-Value Cache (KV Cache) to alleviate computational overhead in subsequent operations. 

The KV Cache is essential for preventing redundant computations in Self-Attention, thereby enhancing inference speed and resource efficiency. For instance, if the Prefill stage computes and stores the Key and Value tensors for the input "I am a," the Decode stage can reuse them to rapidly generate the next token, "man," without redundant computations.

In the Decode stage, new tokens are sequentially generated in an Auto-Regressive manner based on previously generated output tokens. Here, the stored KV Cache is reused to reduce the computational burden of repeated Self-Attention operations and improve processing speed. However, the size of the KV Cache increases significantly with the input length and model size. 

\begin{figure}[!t]
\centering
    \begin{tabular}{@{}c@{}}
        \includegraphics[width=0.48\textwidth, clip]{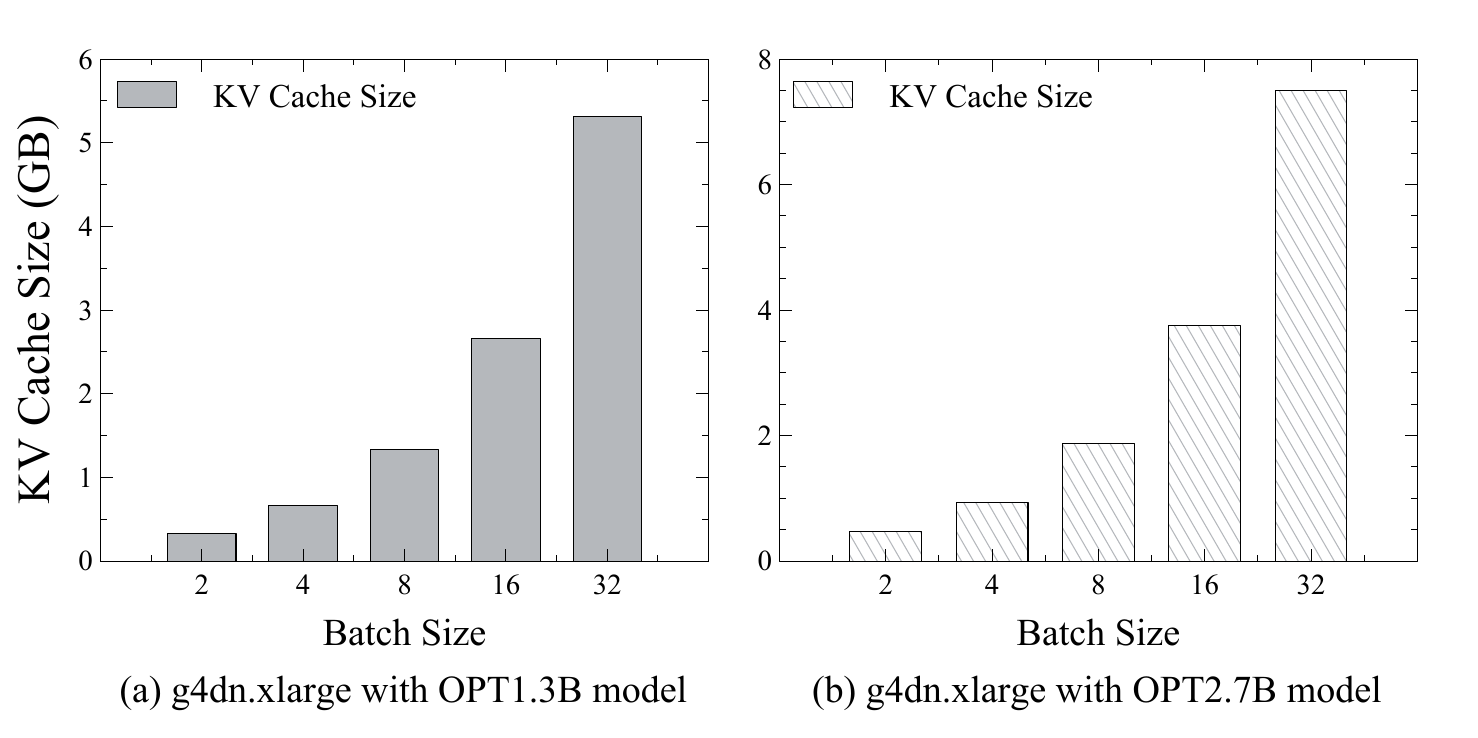}  
        \vspace{-5pt}
    \end{tabular}
    \caption{Analysis of KV Cache size growth across different models in response to increasing batch sizes.
    }
    \vspace{-10pt}
\label{fig:KV Cache Size Diff}
\end{figure}

For example, as shown in Figure~\ref{fig:KV Cache Size Diff}, in the OPT\_2.7B model running on an AWS g4dn.xlarge instance with 1024 input tokens, the KV Cache consumes approximately 0.332GB at a batch size of 2. When the batch size increases to 32, the KV Cache expands to 5.312GB, which can lead to GPU memory exhaustion. This memory constraint may degrade overall system throughput and reduce resource utilization efficiency~\cite{Efficiently2023,vaswani2017attention}.

\subsection{Memory Optimization for LLM Inference via KV Cache Offloading}

During LLM inference, the increasing size of the KV Cache can lead to GPU memory exhaustion, resulting in an Out-of-Memory (OoM) issue. To address this, KV Cache Offloading techniques have been proposed~\cite{flexgen,layerkv,instinferinstorageattentionoffloading,Deepspeed}. These techniques operate by offloading KV Cache data that exceeds GPU memory capacity to CPU memory or disk and retrieving it back to the GPU when needed for computation. This approach effectively alleviates the GPU memory pressure, enabling the processing of long sequences and large batch sizes. Additionally, it allows efficient inference on lower-end GPUs without requiring additional high-performance GPUs, thus reducing deployment costs.

However, latency introduced by data transfer between the GPU and external storage (e.g., CPU memory or disk) is a major limitation of KV Cache Offloading. If the transfer frequency of KV Cache data is high, the increased latency can lead to bandwidth bottlenecks, ultimately degrading inference performance. Therefore, for effective deployment of KV Cache Offloading, it is essential to optimize the process by considering LLM inference characteristics (e.g., sequence length, batch size) and user-defined Service Level Objectives (SLOs), such as maximum allowable response time.
\subsection{Challenges of LLM Inference and KV Cache Offloading in the Cloud}

Cloud service providers (CSPs) such as Amazon AWS offer a variety of GPU virtual machine (VM) instances. As shown in Table~\ref{tbl:awstypes}, the price of these instances varies significantly, ranging from \$0.379 (g4ad.xlarge) to \$40.96 (p4de.24xlarge), depending on the type of GPU, the memory capacity, and the bandwidth of the network~\cite{AWSInstanceTypes}. 

Moreover, when applying KV Cache Offloading to LLM inference, the trade-off between inference performance and actual cost introduces a complex dilemma. To maximize cost-efficiency, users must carefully optimize their choice of VM and offloading strategy based on:
(i) Model size,
(ii) Sequence length, and
(iii) Service Level Objectives (SLOs), such as maximum response time.

However, a systematic framework for making these decisions is currently lacking. As a result, users must experiment with multiple VM options and offloading policies manually to determine an optimal configuration, which adds significant overhead~\cite{flexgen,Deepspeed}.

In this paper, we outline the key dilemmas of KV Cache Offloading for LLM inference in the cloud as follows.

\begin{table}[!t]
\small
\scriptsize 
\caption{\small Various Types of instances provided by AWS.
\\This information was available on Feburary 4, 2025 in N.Virginia region.}

\vspace{-6pt}
\label{tbl:awstypes}
\resizebox{0.48\textwidth}{!}{
    \begin{tabular}{@{}c@{}|c|c|c|c|c|c|c|c@{}}
    \toprule[1.5pt]
    \multirow{2}{*}{Name} & GPU & On- & GPU & FLOPS & vCPU & GPU Mem & Mem & Network
    \\ & Type & Demand (\$) & (\#) & (TFLOPS) & (GiB) & (GiB) & (Gbps) & (Gbps) \\ \hline \hline
    g4dn.xlarge & T4 & \cellcolor{yellow!12}0.526 & \cellcolor{green!5} 1 & 8.141 & 4 & 16 & 16 & - 25  \\ \hline
    g4ad.xlarge & V520 Pro  & \cellcolor{yellow!10}0.379 &\cellcolor{green!5} 1 & 7.373 & 4 & 8 & 16 & - 10    \\ \hline
    g5.xlarge & A10G & \cellcolor{yellow!20}1.006 &\cellcolor{green!5} 1 & 31.52 & 4 & 24 & 16 & - 10    \\ \hline
    g5g.xlarge & T4G  &\cellcolor{yellow!11}0.42 & \cellcolor{green!5} 1 & 8.141 & 4 & 16 & 8 & - 10   \\ \hline
    g6.xlarge & L4  &\cellcolor{yellow!15}0.805 & \cellcolor{green!5} 1 & 30.29 & 4 & 24 & 16 & - 10     \\ \hline
    g6.4xlarge & L4 & \cellcolor{yellow!24}1.323 & \cellcolor{green!5} 1 & 30.29 & 16 & 24 & 64 & - 25   \\ \hline \hline
    
    g4dn.12xlarge & T4 & \cellcolor{yellow!35}3.912 &\cellcolor{green!20} 4 & 8.141 & 48 & 64 & 192 & 50 \\ \hline   
    g4dn.metal & T4 & \cellcolor{yellow!50}7.824 &\cellcolor{green!30} 8 & 8.141 & 96 & 128 & 384 & 100   \\ \hline
    g4ad.16xlarge & V520 Pro & \cellcolor{yellow!30}3.468 &\cellcolor{green!20} 4 & 7.373 & 64 & 32 & 256 & 25  \\ \hline
    g5.12xlarge & A10G & \cellcolor{yellow!40}5.672 & \cellcolor{green!20} 4 & 31.52 & 96 & 96 & 192 & 40 \\ \hline
    g5g.16xlarge & T4G & \cellcolor{yellow!28}2.744 & \cellcolor{green!10} 2 & 8.141 & 64 & 32 & 128 & 25 \\ \hline   
    g6.12xlarge & L4 & \cellcolor{yellow!40}4.602 & \cellcolor{green!20} 4 & 30.29 & 48 & 96 & 192 & 40  \\ \hline
    g6.48xlarge & L4 & \cellcolor{yellow!60}13.35 &\cellcolor{green!30} 8 & 30.29 & 192 & 196 & 768 & 100 \\ \hline
    p4de.24xlarge & A100 & \cellcolor{yellow!90}40.96 &\cellcolor{green!70} 96 & 19.49 & 96 & 7680 & 640 & 400 \\ \bottomrule[1.5pt]
    \end{tabular}
}
\vspace{-15pt}
\end{table}
\squishlist
    \item \textbf{Dual Nature of KV Cache Offloading:} KV Cache Offloading mitigates GPU memory shortage issues, allowing for the processing of larger batch sizes (e.g., greater than 16). However, it increases latency due to data transfer between CPU and GPU (e.g., up to 20\% latency increase in FlexGen~\cite{flexgen}). Specifically, when the sequence length exceeds 4096, the KV Cache size grows significantly (e.g., exceeding 3.2GB), making offloading essential. This, however, increases the likelihood of failing to meet Service Level Objectives (SLOs) such as a 100ms response time.
    \item \textbf{Complexity of Cloud VM Selection:} As shown in Table~\ref{tbl:awstypes}, the performance and cost between instances like g4dn.xlarge (\$0.526, 16\,GiB GPU Memory) and p4de.24xlarge (\$40.96, 7680\,GiB GPU Memory) vary significantly. The optimal VM selection depends on the model requirements (e.g., memory usage, computation speed). High-performance VMs reduce the need for KV Cache Offloading, while lower-end VMs increase reliance on offloading.
    \item \textbf{Difficulty of SLO-Based Optimization:} High-performance VMs (e.g., g6.48xlarge) solve the Out-of-Memory (OoM) problem but may lead to GPU utilization dropping below 50\% when the inference load is low, resulting in wasted costs. On the other hand, lower-end VMs (e.g., g4ad.xlarge) have lower initial costs but suffer from frequent KV Cache Offloading due to VRAM limitations, causing latency to increase by more than double~\cite{Deepspeed}. This results in a dilemma of (i) resource wastage with high-cost VM selection, and (ii) performance degradation with low-cost VM selection.
    \item \textbf{Lack of Automated Optimization Systems:} Currently, there is a lack of guidelines for automating the selection of VMs and KV Cache Offloading in cloud environments. Users must manually test various VMs (e.g., g5 vs. g6 series) and offloading settings, which increases time and cost burdens.
\squishend

This study proposes the necessity of a framework that automatically recommends optimal VM and KV Cache Offloading strategies based on SLO, and introduces a model (Solver) that can balance cost and performance.

\subsection{Existing Approaches and Their Limitations}
\label{subsec:Existing Approaches}
Existing research aiming to optimize LLM inference in cloud environments~\cite{melange, aladdin, splitwise, thunderserve} reveals limitations in achieving cost-efficient LLM serving as they do not consider an integrated approach to KV Cache Offloading and VM selection.

Melange~\cite{melange} proposes an allocation strategy that minimizes cost by mixing different GPU types (e.g., high-performance GPUs and low-cost GPUs) based on LLM service characteristics such as request size, frequency, and Service Level Objectives (SLOs, e.g., response time within 200ms). However, this method relies on profiling GPU performance and the workload pattern in advance, making it difficult to apply to new environments or models. Furthermore, it does not account for KV Cache Offloading, failing to provide optimization solutions in memory-constrained scenarios (e.g., when sequence length exceeds 4096).

Aladdin~\cite{aladdin} suggests a framework for jointly optimizing request batches and resource scaling to meet SLOs. For instance, it adds additional GPUs to reduce latency at high request rates. However, it does not integrate the memory-saving effects of KV Cache Offloading or the trade-offs between different GPU types, which limits the flexibility in VM configuration.

SplitWise~\cite{splitwise} and ThunderServe~\cite{thunderserve} utilize a Phase Splitting strategy, separating the Prefill (initial token generation) and Decode (subsequent token generation) stages. These approaches allocate specialized GPUs to each stage (e.g., high-speed GPUs for Prefill and memory-centric GPUs for Decode) to enhance efficiency. However, this method is only effective in environments where the two stages can be physically separated, making it difficult to apply to standard single-VM-based LLM serving. Additionally, transferring KV Cache between stages requires high-speed interconnects (e.g., NVLink, with GPU-to-GPU bandwidth above 100GB/s), which reduces practicality in cloud VMs without NVLink (e.g., AWS g4dn series).

\index{figure}

Meanwhile, DeepVM~\cite{deepvm}, which deals with deep learning optimization in cloud environments, focuses on optimizing VM clusters for checkpoint-based distributed CNN training. For example, it reduces costs by leveraging saved states during training interruptions. However, this method is tailored for training and is not directly applicable to real-time inference or KV Cache management in LLM serving.

\section{Problem Definition}

\subsection{Definition of Service Level Objective (SLO) Metrics}

In cloud environments, large language model (LLM) inference involves a complex trade-off between memory constraints, cost, and service quality. Depending on the type of inference task, users may have different Service Level Objectives (SLOs).

In this paper, we define two types of inference tasks: Online Inference and Offline Inference.

\begin{itemize}
    \item \textbf{Online Inference} (e.g., chatbots, voice assistants) prioritizes low response latency (e.g., within 100ms) over query throughput, as real-time responsiveness is crucial. Thus, response time is used as the primary SLO metric.

    \item \textbf{Offline Inference} (e.g., batch processing of large datasets) prioritizes high query throughput over response latency, making throughput the primary SLO metric.
\end{itemize}

To encompass both of these metrics under a unified framework, we define Tokens Per Second (TPS) as the SLO metric. TPS represents the number of tokens processed per second, including both input tokens ($L_{\text{in}}$) and output tokens ($L_{\text{out}}$).

LLM inference is typically performed in batches, where a batch consists of multiple queries ($BS$). Given that the total processing time for a batch is denoted as $T_{\text{E2E}}$, TPS is defined as follows:

\begin{equation}
\text{TPS} = \frac{\text{BS} \times \left( L_{\text{in}} + L_{\text{out}} \right)}{T_{\text{E2E}}}
\end{equation}

\subsection{Definition of Cost Efficiency}
In this study, our primary objective is to minimize user costs while ensuring that inference tasks meet their designated SLOs. To achieve this, we define a cost efficiency metric based on the previously introduced Tokens Per Second (TPS) metric.

Let $TPS_{SLO}$ denote the target TPS required by the user to meet the SLO, and let $TPS_{actual}$ represent the actual throughput achieved during inference. Considering that the effective processing rate cannot exceed the user-defined SLO threshold, the effective TPS is defined as: $TPS_{effective} = \min(TPS_{actual}, TPS_{SLO})$

Given this, the total time required to process a batch of queries, denoted as $T_{task}$, is calculated as:

\begin{equation} T_{task} = \frac{{BS} \times ({L_{in}} + {L_{out}})}{TPS_{{effective}} \times 3600} \end{equation}

In cloud environments, GPU usage is typically billed on an hourly basis. Therefore, we apply a ceiling function to $T_{task}$ to account for the actual billable time.

Based on this, we define SLO-based cost efficiency (CE) as a metric to evaluate the cost-effectiveness of a given inference task while ensuring compliance with the SLO. Let VM Price represent the hourly cost of the virtual machine (in dollars per hour). The cost efficiency metric is then defined as:

\begin{equation} CE_{task} = \frac{TPS_{{effective}} \times 3600}{\lceil T_{task} \rceil \times {VM\ Price}} \end{equation}

This metric provides a quantitative measure of how efficiently a system meets the required SLO while optimizing costs in a cloud-based inference environment.

\subsection{Preliminary Results}
\begin{figure}[t!]
    \centering
    \begin{minipage}{0.49\columnwidth}  
        \centering
        \includegraphics[width=\textwidth]{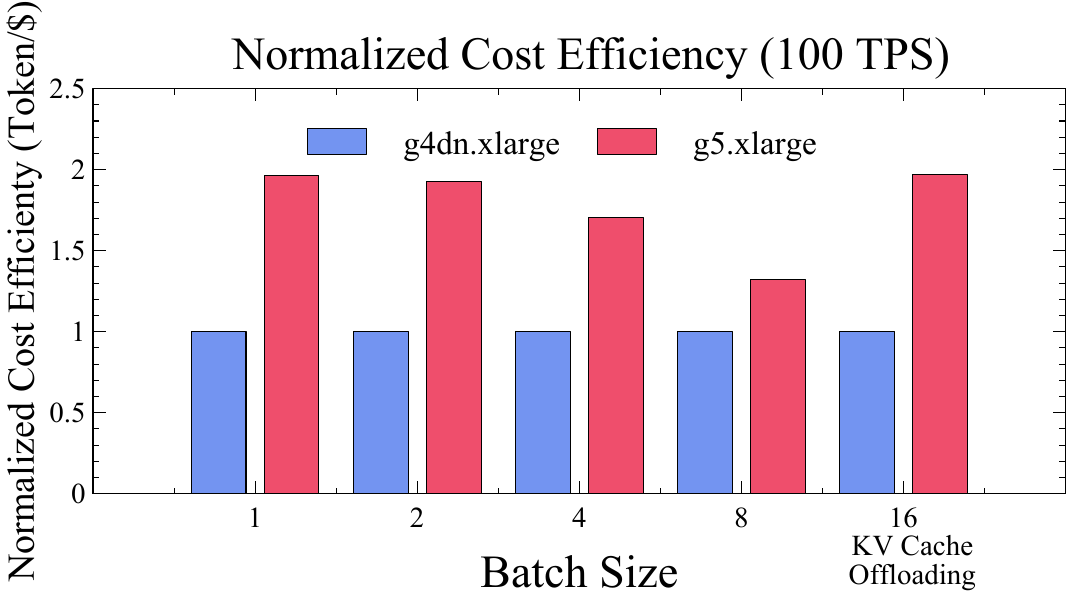}
        \label{fig:tight_graph}
    \end{minipage}
    \hfill
    \begin{minipage}{0.49\columnwidth}  
        \centering
        \includegraphics[width=\textwidth]{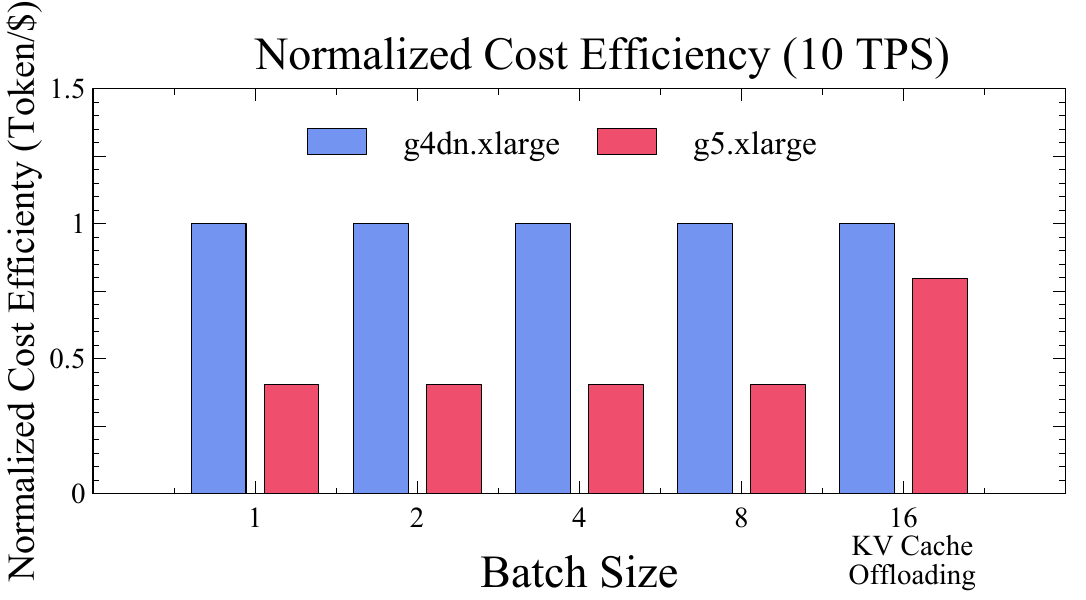}
        \label{fig:loose_graph}
    \end{minipage}
    \vspace{-10pt}
    \caption{Comparison of cost efficiency per GPU instance based on SLO constraints and batch size, based on experimental results using the OPT-2.7B model on an AWS g4dn.xlarge instance with an input length of 512 tokens and an output length of 128 tokens.
    } 
    \vspace{-10pt}
    \label{fig:motivation}
\end{figure}

As shown in Table~\ref{tbl:awstypes} in Section~\ref{sec:background}, cloud VM instances exhibit significant differences in both performance and cost. This variability makes it challenging for users to select the most cost-efficient instance for LLM inference tasks. To validate the complexity of this decision-making process, we evaluated the Cost Efficiency (CE) of two representative VM instances (g4dn.xlarge and g5.xlarge) under different batch sizes and SLO requirements. The experiments were conducted for both cases: with and without KV Cache offloading, assessing its impact on cost efficiency. The results of these experiments are quantitatively presented in Fig.~\ref{fig:motivation}.

\textbf{In a strict SLO environment (100 TPS)}, g5.xlarge demonstrated higher cost efficiency than g4dn.xlarge even at small batch sizes (Batch Size < 16). This is because g5.xlarge delivers higher performance under high-throughput requirements, allowing it to maintain superior cost efficiency over g4dn.xlarge even at smaller batch sizes. At Batch Size 16, g4dn.xlarge faced GPU memory constraints, necessitating KV Cache offloading, which further reduced its cost efficiency. In contrast, g5.xlarge had sufficient memory to operate without offloading, maintaining consistently high cost efficiency as the batch size increased.

\textbf{In a relaxed SLO environment (10 TPS)}, g4dn.xlarge exhibited higher cost efficiency than g5.xlarge at smaller batch sizes (Batch Size < 16). This is because, under relaxed SLO conditions, instance cost became a more critical factor than raw performance. At Batch Size 16, despite g4dn.xlarge requiring KV Cache offloading due to GPU memory limitations, the performance degradation caused by offloading was not a major issue under the relaxed SLO constraints. As a result, g4dn.xlarge, with its lower instance cost, achieved higher cost efficiency compared to g5.xlarge.

To sum up, cost efficiency varies significantly depending on SLO settings and GPU memory utilization strategies, demonstrating that using a high-performance GPU is not always the optimal choice. Particularly in offline inference tasks, where response time constraints are less stringent, KV Cache offloading techniques can enable cost-efficient inference even on lower-cost GPUs. These findings highlight that the optimal GPU instance selection depends on the user’s SLO requirements and the characteristics of the inference task.

\section{Design of \tbdfixed{}}

\subsection{\tbdfixed{}: A Cost-Efficient VM Selection Framework}

Selecting a cost-efficient VM instance in a cloud environment is a challenging task for users. To address this issue, we propose \tbdfixed{}, a software tool designed to assist users in making cost-efficient VM selections. The \tbdfixed{} framework operates in the following four stages:

\begin{enumerate} \setlength{\leftskip}{-10pt}
\item \textbf{Stage 1 Requirement Analysis and Parameter Extraction}: The user provides input parameters, including cost constraints, model characteristics, and performance requirements.

\item \textbf{Stage 2 Resource Suitability Assessment and Candidate Instance Identification}: Based on the provided parameters, the framework calculates the required memory capacity, analyzes the feasibility of KV Cache offloading, and identifies a set of suitable GPU instance candidates.

\item \textbf{Stage 3 Performance-Cost Prediction Modeling}: Leveraging pre-profiled performance data, the framework predicts the TPS of each candidate GPU instance and evaluates its cost efficiency.

\item \textbf{Stage 4 SLO-Based Optimization and Instance Selection}: The framework recommends the most cost-efficient GPU instance that satisfies the SLO constraints.
\end{enumerate}

\begin{table}[!t]
\centering
\small
\caption{Notation and Formulas for Model and Memory Computation}
\label{tab:variables}
\resizebox{0.48\textwidth}{!}{ 
\begin{tabular}{|c|c|}
\hline
\multicolumn{2}{|l|}{\textbf{User Input Parameters}} \\ \hline
\textbf{Variable} & \textbf{Description and Formula} \\ \hline
$BS$ & Batch size \\ \hline
$L_{in}$ & Input token length \\ \hline
$L_{out}$ & Output token length \\ \hline
$P_{\text{max}}$ & User max price willingness \\ \hline
$TPS_{\text{SLO}}$ & User SLO Requirement\\ \hline
\hline
\multicolumn{2}{|l|}{\textbf{Model Parameters}} \\ \hline
\textbf{Variable} & \textbf{Description and Formula} \\ \hline
$h_1$ & Hidden Size (model dimension) \\ \hline
$h_2$ & Intermediate Size (projection) \\ \hline
$nh$ & Number of Attention Heads \\ \hline
$L$ & Transformer layers \\ \hline
$C_{\text{off}}$ & KV cache offloading ratio \\ \hline
$\text{Precision}_{\text{bytes}}$ & Bytes per parameter (e.g., FP16 = 2B) \\ \hline
$\text{Mem}_{\text{model}}$ (Model Size) & Number of Model Parameters $\cdot \text{Precision}_{\text{bytes}}$ \\ \hline
$\text{Mem}_{\text{KVcache}}$ (KV Cache Size) & $2 \cdot BS\cdot (L_{in} + L_{out}) \cdot nh \cdot \text{Precision}_\text{bytes} \cdot L$ \\ \hline
$\text{Mem}_{\text{KVcache, per\_layer}}$ & KV Cache per layer: $\frac{\text{Mem}_{\text{KVcache}}}{L}$ \\ \hline
$\text{Mem}_{\text{activation}}$(Activation Size) & $2\cdot(L_{in} + L_{out})\cdot BS \cdot h_1$ \\ \hline
\hline
\multicolumn{2}{|l|}{\textbf{Instance Specifications}} \\ \hline
\textbf{Variable} & \textbf{Description and Formula} \\ \hline
$FLOPS_\text{GPU}$ & GPU's theoretical FLOPS \\ \hline
$\text{BW}_{\text{gpu} \to \text{cpu}}$ & Bandwidth for GPU-to-CPU data transfer \\ \hline
$\text{BW}_{\text{cpu} \to \text{gpu}}$ & Bandwidth for CPU-to-GPU data transfer \\ \hline
\end{tabular}
} 
\end{table}

\subsection{Requirement Analysis and Parameter Extraction}

This stage involves collecting key input parameters necessary for LLM inference tasks. The most critical parameter is the maximum willingness-to-pay price ($P_{\text{max}}$), which represents the maximum cost (\$/hour) that the user is willing to pay. This value serves as a fundamental constraint in the subsequent stages of the algorithm, determining the range of GPU instances that can be considered.

Additionally, the user specifies the target LLM model (e.g., OPT-2.7B, LLaMA-7B), and based on this selection, the system automatically extracts key model parameters such as model size, number of attention heads, head dimensions, feed-forward network (FFN) dimensions, and activation size. Other essential input parameters include the average input token length, average output token length, batch size, and the required SLO in terms of TPS (Tokens Per Second).

This stage plays a crucial role in transforming user requirements into quantitative parameters, establishing the foundation for resource suitability assessment and performance prediction. Ultimately, it is essential for selecting the most cost-efficient GPU instance that meets both performance objectives and budget constraints.

\subsection{Resource Suitability Assessment and Candidate Instance Identification}
At this stage, the system evaluates the memory requirements for inference based on the collected user parameters and assesses the feasibility of KV Cache Offloading to identify the most suitable GPU instance candidates.
First, the system calculates the total memory requirement \(\text{Mem}_{\text{total}} \) for the given Transformer-based LLM model and its input-output parameters. This is defined as the sum of the following three components:
\( \text{Mem}_{\text{total}} = \text{Mem}_{\text{model}} + \text{Mem}_{\text{activation}} + \text{Mem}_{\text{KVcache}} \).  
Additionally, the base memory requirement is defined as: \( \text{Mem}_{\text{base}} = \text{Mem}_{\text{model}} + \text{Mem}_{\text{activation}} \).
These stages follow three key criteria to evaluate GPU instance suitability and Algorithm~\ref{alg:resource-evaluation-price-first}.

\noindent\textbf{Case1) No Offloading Required:}  
If the available GPU memory is greater than or equal to the total memory requirement, i.e., \( \text{GPU}_{\text{memory}}^i \geq \text{Mem}_{\text{total}} \) then the instance can fully accommodate the model without KV Cache Offloading. Here, $i$ refers to the current particular running instance. In this case, the offloading coefficient is set to \( C_{\text{off}}^i = 0 \) and the instance is added to the candidate pool.

\noindent\textbf{Case2) Offloading Not Feasible:}  
An instance is deemed unsuitable if it meets any of the following conditions:
\begin{itemize}
    \item If the available GPU memory is smaller than the model weights:  
    \quad \( \text{GPU}_{\text{memory}}^i < \text{Mem}_{\text{model}} \).

    \item If the KV Cache size per layer exceeds the available memory:
    \quad \( \text{Mem}_{\text{KVcache, per\_layer}} > \text{Mem}_{\text{avail}}^i \).  
\end{itemize}
This condition arises because attention operations are performed on the GPU, requiring KV Cache to remain in GPU memory.
When the available memory is insufficient, an Out of Memory (OOM) error occurs, preventing execution.

\noindent\textbf{Case3) KV Cache Offloading Required:}  
If an instance does not fall into either of the previous categories, KV Cache Offloading is required. In this case, the offloading coefficient is computed as:  
\( C_{\text{off}}^i = 1 - \frac{\text{Mem}_{\text{avail}}^i}{\text{Mem}_{\text{KVcache}}} \)

\begin{algorithm}[!t]
\caption{Resource Suitability Evaluation and Instance Selection (Price Priority)}
\label{alg:resource-evaluation-price-first}
\SetAlgoLined
\scriptsize
\KwIn{
    \textbf{Memory Requirements:} \\
    $\text{Mem}_{\text{model}}$ — Model weight memory requirement \\
    $\text{Mem}_{\text{activation}}$ — Activation memory requirement \\
    $\text{Mem}_{\text{KVcache}}$ — Total KV Cache memory requirement \\
    $\text{Mem}_{\text{KVcache, per\_layer}}$ — KV Cache memory per layer \\
    \textbf{For each GPU instance $i$:} \\
    $\text{GPU}_{\text{memory}}^i$ — Total GPU memory \\
    $\text{GPU}_{\text{price}}^i$ — GPU price \\
    \textbf{User-defined maximum price:} $P_{\text{max}}$
}

\KwOut{GPU candidates that satisfy both price and resource conditions}

$ \text{Candidates} \gets \emptyset$ \tcp{Initialize candidate set}

$ \text{Mem}_{\text{total}} \gets \text{Mem}_{\text{model}} + \text{Mem}_{\text{activation}} + \text{Mem}_{\text{KVcache}}$ \;
$ \text{Mem}_{\text{base}} \gets \text{Mem}_{\text{model}} + \text{Mem}_{\text{activation}}$ \;

\For{each GPU instance $i$}{
    \If{$\text{GPU}_{\text{price}}^i \leq P_{\text{max}}$}{ \tcp{Apply Price Filter}
        $ \text{Mem}_{\text{avail}}^i \gets \text{GPU}_{\text{memory}}^i - \text{Mem}_{\text{base}}$ \tcp{Calculate Available Memory}
        
        \eIf{$\text{GPU}_{\text{memory}}^i \geq \text{Mem}_{\text{total}}$}{ \tcp{Offloading Not Required}
            $ C_{\text{off}}^i \gets 0$ \;
            Add $(i, C_{\text{off}}^i)$ to Candidate Set \;
        }{
            \If{$\text{GPU}_{\text{memory}}^i < \text{Mem}_{\text{model}}$ \textbf{OR} $\text{Mem}_{\text{KVcache, per\_layer}} > \text{Mem}_{\text{avail}}^i$}{ \tcp{Offloading Not Possible}
                Mark GPU $i$ as Unsuitable \tcp{Exclude from candidates}
            }
            \Else{ \tcp{KV Cache Offloading Required}
                $ C_{\text{off}}^i \gets 1 - \frac{\text{Mem}_{\text{avail}}^i}{\text{Mem}_{\text{KVcache}}}$ \;
                \If{$\text{Mem}_{\text{KVcache, per\_layer}} \leq \text{Mem}_{\text{avail}}^i$}{ \tcp{Layer-Level Constraint Check}
                    Add $(i, C_{\text{off}}^i)$ to Candidate Set \;
                }
                \Else{
                    Mark GPU $i$ as Unsuitable \tcp{Exclude from candidates}
                }
            }
        }
    }
}

Sort Candidate Set by $\text{GPU}_{\text{price}}^i$ in ascending order \;
\Return Candidate Set $\text{Candidates}$ \;
\end{algorithm}

Finally, the selected instances are sorted in ascending order based on cost, and the results are used as input for the performance-cost prediction modeling stage. This systematic approach ensures that the most cost-efficient GPU instance is selected within the user's budget while accurately evaluating the feasibility and cost-efficiency of KV Cache Offloading.

\subsection{Instance Performance Prediction}

At this stage, the system predicts Tokens Per Second (TPS) for the candidate GPU instances identified in the previous step. This is achieved through mathematical modeling that leverages model parameters, hardware profiling information (FLOPS, bandwidth, etc.) of each candidate instance, and the offloading coefficient to quantitatively estimate the task processing time.

The total task processing time \(T_{\text{task}}\) consists of the Prefill and Decode stages and is calculated as follows~\cite{flexgen}:
\begin{itemize}
    \item \textbf{Prefill Stage:} This stage processes the entire input sequence. The processing time per layer (\(T_{\text{pre}}\)) is multiplied by the number of layers (\(n\)).
    \item \textbf{Decode Stage:} This stage generates each output token sequentially. The processing time per layer (\(T_{\text{dec}}\)) is multiplied by the number of layers (\(n\)) and the number of generated tokens (\(L_{\text{out}}-1\)), since the first output token is already processed in the Prefill stage.
\end{itemize}
Thus, the total task processing time \(T_{\text{task}}\) is expressed as follows:
\begin{equation}
T_{\text{task}} =\underbrace{ T_{\text{pre}} \cdot n}_\text{Prefill Time} + \underbrace{T_{\text{dec}} \cdot n\cdot (L_\text{out}-1)}_\text{Decode Time}    
\end{equation}

\paragraph{Prefill Stage Processing Time}
The Prefill stage processing time \(T_{\text{pre}}\) consists of computation time and KV Cache storage time. Since GPU computation and KV Cache offloading occur in parallel, the total delay is determined by the process with the longest execution time:
\begin{equation}
T_{\text{pre}} = \max\left( CTCF(T^\text{p}_\text{compute}), T_{\text{trans}}^{\text{p}} \right)
\end{equation}
The computation time \(T_{\text{compute}}^p\) in the Prefill stage is divided into Linear Layer computation and Self-Attention computation, both of which are calculated by dividing the required floating-point operations (FLOPs) by the GPU's theoretical FLOPS capacity:
\[
T_{\text{compute}}^p =
\underbrace{\frac{BS\cdot(8L_\text{in}\cdot h_1^2 + 4L_\text{in}\cdot h_1 \cdot h_2)}{FLOPS_{\text{GPU}}}}_{\text{Linear Layer compute time}} + 
\underbrace{\frac{4 \cdot BS\cdot L_\text{in}^2 \cdot h_1}{FLOPS_{\text{GPU}}}}_{\text{Self-Attention compute time}}
\]
The KV Cache transfer time \(T_{\text{trans}}^p\) in the Prefill stage represents the time required to offload the generated KV Cache from GPU to CPU memory and is computed as:
\[
T_{\text{trans}}^p = \frac{C_\text{off}\cdot\{2\cdot (L_\text{in}+1)\cdot h_1\cdot \text{Precision}_\text{bytes}\}\cdot BS}{\text{BW}_{\text{gpu} \to \text{cpu}}}
\]

\paragraph{Decode Stage Processing Time}
The Decode stage processing time \(T_{\text{dec}}\) includes computation time and KV Cache retrieval time. If the KV Cache fully resides in the GPU, only computation time is considered. However, if offloading occurs, additional latency is introduced due to data transfer from CPU to GPU. The Decode time is therefore expressed as:
\begin{equation}
T_{\text{dec}} = CTCF(T^\text{d}_\text{compute})+ T_{\text{trans}}^{\text{d}}
\end{equation}

The Decode computation time \(T_{\text{compute}}^d\) consists of Linear Layer and Self-Attention computation, and is computed as follows:
\[
T_{\text{compute}}^{d} = \underbrace{\frac{BS\cdot(8h_1^2 +  4h_1 \cdot h_2)}{FLOPS_\text{GPU}}}_{\text{Linear Layer compute time}} + \underbrace{\frac{4 \cdot BS\cdot (L_\text{in} + \frac{L_\text{out}}{2}) \cdot h_1}{FLOPS_\text{GPU}}}_\text{Self-Attention compute time}
\]
The KV Cache transfer time \(T_{\text{trans}}^d\) in the Decode stage refers to the time required to load KV Cache stored in CPU memory back into GPU memory and is computed as:
\[
T_{\text{trans}}^d = \frac{\left( C_{\text{off}} \cdot 2 \cdot (L_{\text{in}}+1) + L_\text{out} \right) \cdot h_1 \cdot \text{Precision}_{\text{bytes}} \cdot BS}{\text{BW}_{\text{cpu} \to \text{gpu}}}
\]

This modeling approach accounts for both cases where offloading is necessary and unnecessary, effectively considering GPU memory constraints and computational performance. By incorporating both computation latency and KV Cache offloading overhead, this approach enables a quantitative analysis of the trade-off between computation and memory access time in both Prefill and Decode stages.

Using this modeling framework, Tokens Per Second (TPS) can be estimated, allowing for the selection of the most optimal GPU instance for a given inference task. While this theoretical modeling provides a solid foundation, it is important to note that GPU manufacturers' theoretical FLOPS values do not always accurately reflect real-world LLM inference workloads. The limitations of this approach, along with the Compute Time Calibration Function (CTCF) designed to correct these discrepancies, are discussed in Section~\ref{CTCF}.

\subsection{Step 4: Final Instance Selection Based on SLO}

Based on the TPS (Tokens Per Second) values computed for each GPU instance in the previous stage, this step selects the most cost-efficient instance while ensuring that the user's Service Level Objective (SLO) is met. The selection process follows these steps:

First, instances that fail to satisfy the user-defined SLO constraint (\(\text{TPS} \geq \text{TPS}_{\text{SLO}}\)) are eliminated from consideration. Next, the cost efficiency metric (Equation 3) is calculated for each remaining instance. Finally, the instance with the highest cost efficiency is selected. In the event of a tie, the instance with the higher TPS is prioritized.

The final selection result is presented to the user along with comprehensive details, including instance type, expected TPS, cost, and KV Cache offloading configuration. Additionally, the system provides alternative options and a performance-cost trade-off analysis, enabling users to make an informed decision that is optimized for their specific LLM inference workload.

\subsection{Compute Time Calibration Function (CTCF)}
\label{CTCF}

\begin{figure}[!t]
\centering
    \begin{tabular}{@{}c@{}}
        \includegraphics[width=0.48\textwidth, clip]{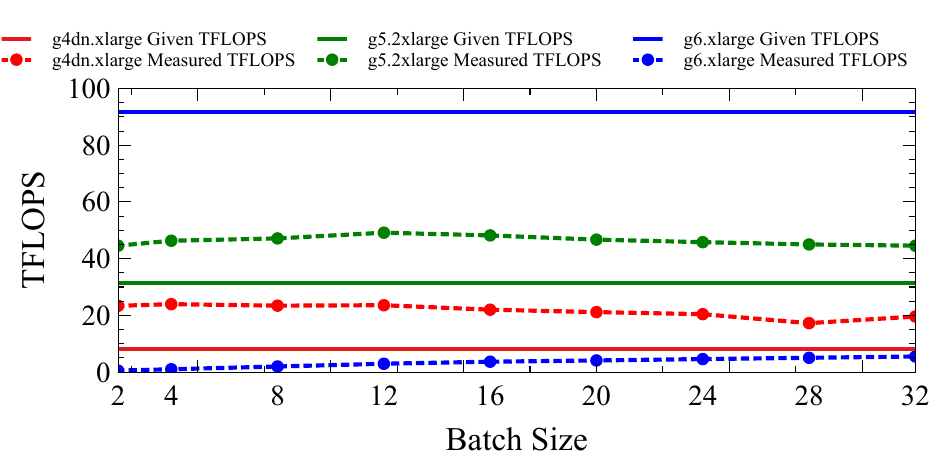}  
        \vspace{-5pt}
    \end{tabular}
    \caption{Comparison of FLOPS provided by the GPU manufacturer (NVIDIA) and the actual FLOPS utilized when calculating Prefill time on AWS GPU VMs. The results present TFLOPS measurements for three different GPU VMs using the OPT-2.7B model with an input size of 512 tokens and an output size of 128 tokens as batch size grows.
    }
    \vspace{-10pt}
\label{fig:flops_diff}
\end{figure}
\index{figure}

The theoretical FLOPS values provided by GPU manufacturers do not accurately reflect real-world performance in LLM inference workloads. Figure~\ref{fig:flops_diff} illustrates the discrepancy between the FLOPS values advertised by the manufacturer and those actually utilized in computation across three different GPU instances. This discrepancy arises from factors such as memory bottlenecks, reduced GPU utilization, and variations in computation patterns, which manifest differently in the Prefill and Decode stages of LLM inference. As a result, selecting a GPU instance solely based on theoretical FLOPS can lead to significant performance mismatches, causing users to incur unnecessary costs. To address this issue, it is essential to introduce a calibration method that aligns theoretical FLOPS values with actual computational performance.

\begin{itemize}
    \item \textbf{CTCF Modeling}: This study conducted preliminary experiments across various batch sizes to analyze the relationship between LLM inference time and batch size. The results consistently showed a linear increase in inference time for both the Prefill and Decode stages. This linear trend was observed across different GPU architectures, including T4, A10G, L4, and L40s, leading to the introduction of a regression-based CTCF model.
\end{itemize}

CTCF is a linear transformation function that adjusts theoretical computation time to match actual execution time. It is defined as follows:
\begin{equation}
CTCF(T_{\text{compute}}) = \alpha \cdot T_{\text{compute}} + \beta
\end{equation}
where $\alpha$ is a scaling factor that corrects overestimation or underestimation of theoretical computation time, and $\beta$ is a fixed offset that compensates for systematic delays caused by GPU execution bottlenecks, memory access latency, and other hardware constraints. These parameters are optimized using the least squares method and are determined through pre-profiling experiments. 

Through extensive pre-profiling, $\alpha$ and $\beta$ values were computed for all AWS GPU instances across various batch sizes and stored as reference data. As shown in Table~\ref{tbl:CTCF_reg}, applying these per-instance $\alpha$ and $\beta$ values significantly reduces the prediction error, bringing the adjusted execution time very close to the actual measurement. Based on this, \tbdfixed{} profiles $\alpha$ and $\beta$ values for all available AWS GPU instances, enabling precise FLOPS-based execution time predictions and recommending the optimal instance for users.

\begin{table}[h!] 
\footnotesize
    \vspace{-5pt}
    \centering
    \caption{\small Values of $\alpha$, $\beta$ to calculate adjusted $T_{Prefill}$ (Model: OPT 2.7B)
    }
    \vspace{-6pt}
    \label{tbl:CTCF_reg}
    \resizebox{0.48\textwidth}{!}
    { \footnotesize
       \begin{tabular}{|l|l|l|l|}
\hline
\textbf{Instance Type (GPU Model)}& \textbf{$\alpha$} & \textbf{$\beta$} & \textbf{avg. error rate ($\%$)} \\ 
\hline
g4dn.xlarge (T4) & -0.185 & 24.35 & 4.47 \\ \hline
g5.2xlarge (A10G) & -0.074 & 46.97 & 2.60 \\ \hline
g6.xlarge (L4) & -0.1238 & 42.52 & 2.23 \\ \hline
\end{tabular}
}
\end{table}

The CTCF-based correction method effectively compensates for the inherent limitations of theoretical FLOPS values provided by GPU manufacturers, leading to more accurate LLM inference performance predictions.

\section{Implementation}
We developed \tbdfixed{} using Python (3.10.14). 
For performance modeling and KV cache offloading optimization, we utilized NumPy (1.24.3) for efficient numerical computations and statistical analysis. Our system is built on top of FlexGen, a cutting-edge framework for LLM inference that provides robust KV cache offloading capabilities.
A key advantage of \tbdfixed{} is its minimal computational overhead and exceptional speed in determining optimal resource configurations. 
Once user parameters and SLO requirements are provided, our system quickly performs TPS predictions and cost-efficiency calculations, enabling rapid and precise GPU instance recommendations.
The complete source code of \tbdfixed{}, along with all associated tools and algorithms, is publicly available for download at
\cb{https://github.com/lass-lab/InferSave}.

\section{Evaluation}

\label{sec:eval}
\subsection{Experimental setup}

For our evaluation, we conducted two contrasting inference tasks representative of online and offline inference scenarios to comprehensively assess the impact of offloading strategies on cost and performance across various cloud-based GPU instances. The objective of the evaluation is to quantitatively analyze the effects of offloading and the impacts it has on cost and performance efficiency, as well as to pick the optimal instance given a SLO as input. Online inferencing focuses on finding the most price-effective inference while meeting the strict SLO requirement, while offline inferencing relaxes the SLO requirement, allowing for strategies such as offloading and used lower priced instances. All experiments were performed 3 times for each instance to maintain result integrity, and the average of each result were used for analysis.

\noindent\textbf{Workload Definition:}
For a holistic evaluation of \tbdfixed{}'s ability to select the optimal instance in a variety of scenarios, we perform two contrasting inference workloads.

\squishlist
    \item \textbf{Online Inference workload:} To model a real-time chatbot system, we use a pattern of 128 input tokens and a 512 output tokens. This simulates a common AI LLM chatbot scenario of a user asking short questions, with the chatbot providing detailed answers. The workload evaluates a total of 3000 requests.

    \item \textbf{Offline Inference workload:} To model a batch processing task, an input size of 1024 tokens and an output size of 128 tokens was used. This takes into account tasks such as document summarization and data wrangling. To simulate a batch processing task, the workload evaluates the performance of completing 1000 requests.
    
\squishend

\noindent\textbf{AWS Cloud Experiment Setup}:
To maintain uniform experimental conditions and reduce potential disruptions caused by fluctuating cloud workloads, all experiments were carried out on AWS in the us-east-1 (Northern Virginia) region between 9:00 AM and 10:00 PM KST, spanning the period from December 2024 to March 2025. To avoid performance variations due to regional resource contention, testing was evenly distributed across availability zones us-east-1a through us-east-1f. For the GPU-VMs, we utilized $\mathtt{g4dn.xlarge}$(NVIDIA T4), $\mathtt{g5.xlarge}$(NVIDIA A10G), $\mathtt{g6.xlarge}$(NVIDIA L4) and $\mathtt{g6e.xlarge}$(NVIDIA L40s)
A detailed specification of the instances are specified in Table ~\ref{tbl:GPU_instances_modified}.

\begin{table}[h]
\centering
\caption{\small Specifications of VM instances, including 4 GPU-VMs based on AWS specifications.}
\vspace{-6pt}
\renewcommand{\tabcolsep}{2mm}
\resizebox{0.5\textwidth}{!}{
\begin{tabular}{|c|c|c|c|c|c|}
\hline
\multirow{2}{*}{\textbf{Instance}} & \multirow{2}{*}{\textbf{GPU-Type}} & \textbf{On-Demand Price } & \textbf{GPU Memory} & \textbf{FP16} & \textbf{PCIe B/W} \\ 
\textbf{} & \textbf{} & \textbf{(\$/hr)} & \textbf{(GB)} & \textbf{(TFLOPS)} & \textbf{(GB/s)} \\ \hline\hline
g6e.xlarge   & L40s & 2.699  & 48 & 91.61 & 12 \\ \hline
g6.xlarge    & L4  & 1.167 & 24 & 30.29 & 12 \\ \hline
g5.xlarge            & A10G  & 1.466  & 24 & 31.52 & 12 \\ \hline
g4dn.xlarge            & T4  & 0.71 & 16 & 8.24 & 6 \\ \hline
\end{tabular}
}
\label{tbl:GPU_instances_modified}
\end{table}

To validate the effectiveness of \tbdfixed{}, major transformer-based LLM models such as OPT-1.3B, OPT-2.7B, OPT-6.7B were used for testing in an in-house benchmark suite. To find the optimal performance configuration, tests were conducted by varying the batch size from 1 to 64 under different conditions for single GPU processing.

\noindent\textbf{Policy To Select Instance}:
As stated in Section~\ref{subsec:Existing Approaches}, there are no clear state of the art methodologies for GPU instance selection for inferencing. Therefore, in our evaluation, we compared the following two baseline approaches with \tbdfixed{}.
\squishlist
    \item \textbf{Most powerful instance(Max-Performance)}: 
    This policy simply chooses the GPU instance that offers the most performance, and aims to lower latency and raise throughput as much as possible. However, this methodology does not take into consideration price, and therefore running costs can be raised needlessly.
    \item \textbf{Simple performance prediction(\tbdfixed{} (without KV Cache offloading))}: 
    This policy uses theoretical performance metrics (FLOPS, memory bandwidth) to predict performance and select an instance. However, it does not take into consideration the effects of KV Cache offloading, and may not be able to find the most optimal instance.  
\squishend

\subsection{CTCF Validation}
\tbdfixed{} proposes the Compute Time Calibration Function (CTCF) to accurately determine the optimal instance based on user requirements. To validate the accuracy of CTCF, experiments were conducted on two GPU instances, g4dn.xlarge and g6.xlarge. The experiments utilized the OPT-2.7B model, with an input token length of 512 and an output token length of 128. The model's key computational units, including a hidden size of 2560 and an intermediate size of \(2560 \times 4\), were applied, and the total number of layers (32) was incorporated to measure computation time. For FLOPS estimation, the theoretical FLOPS values provided by GPU manufacturers were used: g4dn.xlarge with NVIDIA T4 (8.24 TFLOPS) and g6.xlarge with NVIDIA L4 (30.29 TFLOPS).

\begin{figure}[t!]
    \centering
    \begin{tabular}{@{}c@{}}
        \includegraphics[width=0.5\textwidth, clip]{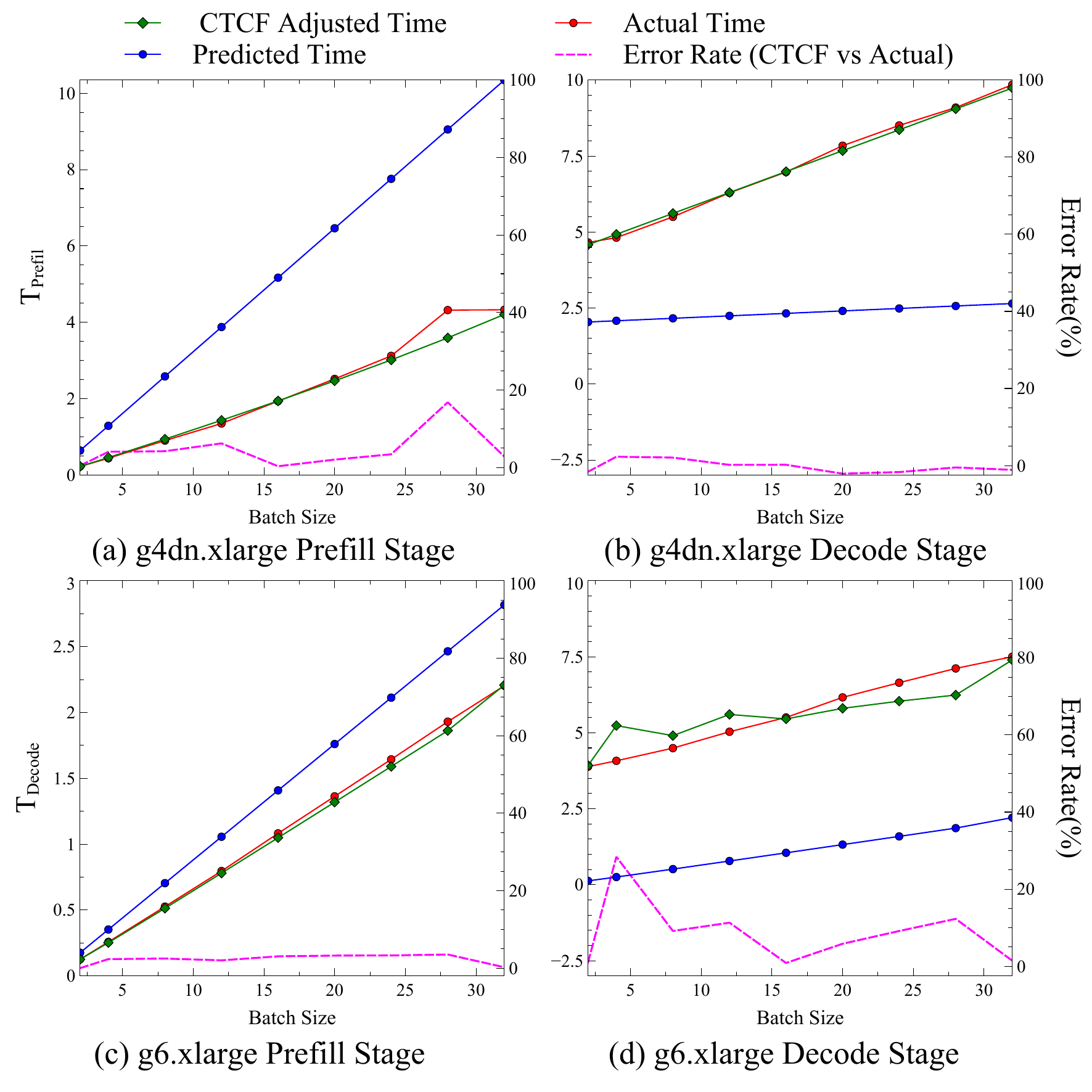}  
        \vspace{-5pt}
    \end{tabular}
    \vspace{-10pt}
    \caption{CTCF accuracy analysis. The results illustrate the predicted time (blue), actual time (red), and CTCF-adjusted values (green) for Prefill and Decode times as batch size increases on two different GPU VMs. Additionally, the Error Rate between the CTCF-adjusted time and actual time is presented.}
    \vspace{-10pt}
    \label{fig:prediction_plot}
\end{figure}

After applying CTCF, the corrected prediction times were computed and compared with actual measurements to analyze the error rate. As shown in Figure~\ref{fig:prediction_plot}, the CTCF-adjusted values closely matched the actual measurements. Specifically, in the Decode stage of g4dn.xlarge, the corrected values exhibited an average error rate of 1\% compared to actual measurements, while in the Prefill stage of g6.xlarge, the average error rate was 2\%. These results demonstrate that the CTCF-adjusted computation time aligns well with real-world measurements, thereby verifying that \tbdfixed{} can accurately recommend the most suitable GPU instance for users.

\subsection{Evaluation results}
To evaluate the effectiveness of \tbdfixed{} and our proposed methodologies, we conducted experiments on both online and offline workloads. While we have performed comprehensive experiments across various model sizes and batch sizes, we have decided to focus on the analysis of representative results using the OPT-2.7B model with a batch size of 32.
This configuration was chosen as it clearly shows the performance variations of each GPU instance, and also demonstrates a good middle ground of performance and resource utilization. We set the maximum cost per hour ($P_{\text{max}}$) to \$3.00/hr. This value was chosen as g6e.xlarge, the most powerful instance in our experiments, has an on-demand cost of \$2.699/hr, and a slightly higher cost than this allows for a fair comparison across all instances. 

\begin{table}[h]
    \centering
    \caption{Comparison of Instance Selection Results by SLO Constraints \\(400 TPS and 600 TPS)
    }
    \label{tab:online-policy_comparison}
    \small 
    \resizebox{\columnwidth}{!}{ 
    \begin{tabular}{c|c|c|c|c}
        \toprule
        \textbf{SLO} & \textbf{Evaluated Policies} & \textbf{Selected Instances} & \textbf{TPS(avg.)} & \textbf{Total Cost(\$)}  \\
        \midrule
        \multirow{3}{*}{400 TPS} & \tbdfixed{}-1st & g4dn.xlarge & 620.17 & 0.71  \\
        & \tbdfixed{}-2nd & g6.xlarge & 802.19 & 1.167\\
        & Max-Performance & g6e.xlarge & 1506.54 & 2.699  \\
        \midrule
        \multirow{3}{*}{600 TPS} & \tbdfixed{}-1st & g6.xlarge & 800.15 & 1.167  \\
        & \tbdfixed{}-2nd & g5.xlarge & 1206.12 & 1.466  \\
        & Max-Performance & g6e.xlarge  & 1505.37 & 2.699  \\
        \bottomrule
    \end{tabular}
    }
\end{table}

\subsubsection{Online inference workload results}
\begin{figure}[!t]
\centering
\begin{minipage}{0.49\columnwidth} 
    \centering
    \includegraphics[width=0.95\textwidth, keepaspectratio]{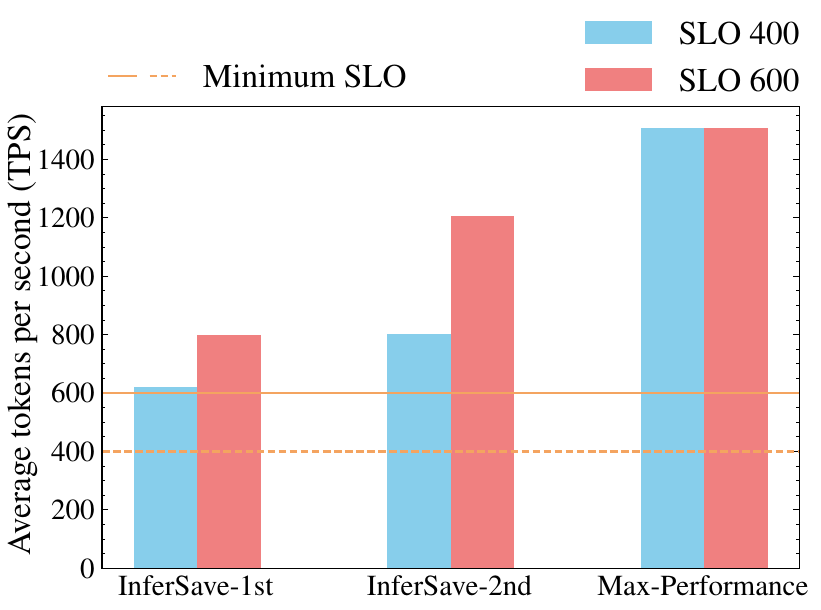}
\end{minipage}
\hfill
\begin{minipage}{0.49\columnwidth} 
    \centering
    \includegraphics[width=0.95\textwidth, keepaspectratio]{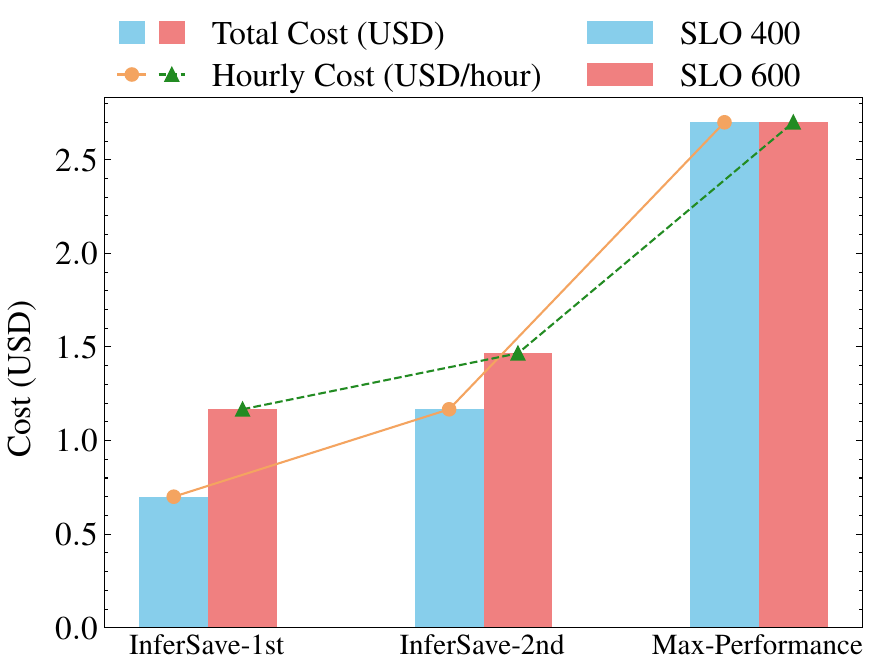}
\end{minipage}

\caption{Comparison of average TPS and cost for different \tbdfixed{} configurations and the baseline configuration under varying SLO constraints for online inference workloads (Left: Average TPS, Right: Cost).}
\vspace{-10pt}
\label{fig:on-line TPS}
\end{figure}

Table \ref{tab:online-policy_comparison} and Figure~\ref{fig:on-line TPS} shows the instances selected by each policy based on the SLO requirements given for an online inference workload, as well as the performance and price comparisons. We analyze the first and second selections of \tbdfixed{}'s policy within two minimum TPS requirements (400 TPS, 600 TPS), and compare it with the selection of the Max-Performance policy's selection. Note that the results of \tbdfixed{} without KV Cache offloading were the same as Max-Performance's selection, and thus were excluded from Table \ref{tab:online-policy_comparison}. 
This result was observed as the workload size used in this experiment was sufficiently small, allowing all KV Cache data to be accommodated within the GPU memory. Therefore, offloading had no impact on performance, and consequently, there was no difference in the selected instances. Additionally, for these experiments, the total runtime did not surpass an hour, leading to the hourly cost and the total cost to be the same.

With an SLO requirement of 400 TPS, \tbdfixed{} selected g4dn.xlarge as its first choice, and this instance offered the lowest cost of \$0.71 while providing 620.17 TPS. On the other hand, Max-Performance selected g6e.xlarge, which provides the highest performance of 1506.54 TPS, but at a cost of \$2.699, which is about 280\% more expensive than \tbdfixed{}'s top choice. A similar pattern was observed with the 600 TPS SLO constraint, with \tbdfixed{}'s selection of g6.xlarge meeting the SLO at a 56.75\% lower cost than g6e.xlarge.

This shows that the instances chosen by the Max-Performance policy overshoots the given SLO requirement greatly, leading to wasted GPU utilization and higher running costs. Meanwhile, \tbdfixed{} demonstrates optimal instance selection by using accurate performance prediction to select the most favorable instance for the given requirements.


\begin{table}
    \centering
    \caption{Comparison of Instance Selection Results by SLO Constraints \\(100 TPS and 200 TPS)}
    \label{tab:offline-policy_comparison}
    \small 
    \resizebox{\columnwidth}{!}{ 
    \begin{tabular}{c|c|c|c|c|c}
        \toprule
        \textbf{SLO} & \textbf{Evaluated Policies } & \textbf{Selected Instances} &\textbf{$C_\text{off}$($\%$)} &\textbf{TPS(avg.)} & \textbf{Total Price(\$)}  \\
        \midrule
        \multirow{3}{*}{100 TPS} & \tbdfixed{}-1st & g4dn.xlarge & 100 &169.17 & 2.13  \\
        & \tbdfixed{}-2nd & g6.xlarge & 60 & 415.04 & 2.344\\
        & Max-Perf., \tbdfixed{}(w/o KV) & g6e.xlarge & 0 & 1506.54 & 2.699  \\
        \midrule
        \multirow{3}{*}{200 TPS} & \tbdfixed{}-1st & g6.xlarge&60 & 414.28 & 2.334  \\
        & \tbdfixed{}-2nd & g5.xlarge& 60 & 414.01 & 2.932  \\
        & Max-Perf., \tbdfixed{}(w/o KV) & g6e.xlarge& 0 & 1505.37 & 2.699  \\
        \bottomrule
    \end{tabular}
    }
\end{table}

\begin{figure}[h]
\centering
\begin{minipage}{0.49\columnwidth} 
    \centering
    \includegraphics[width=\textwidth]{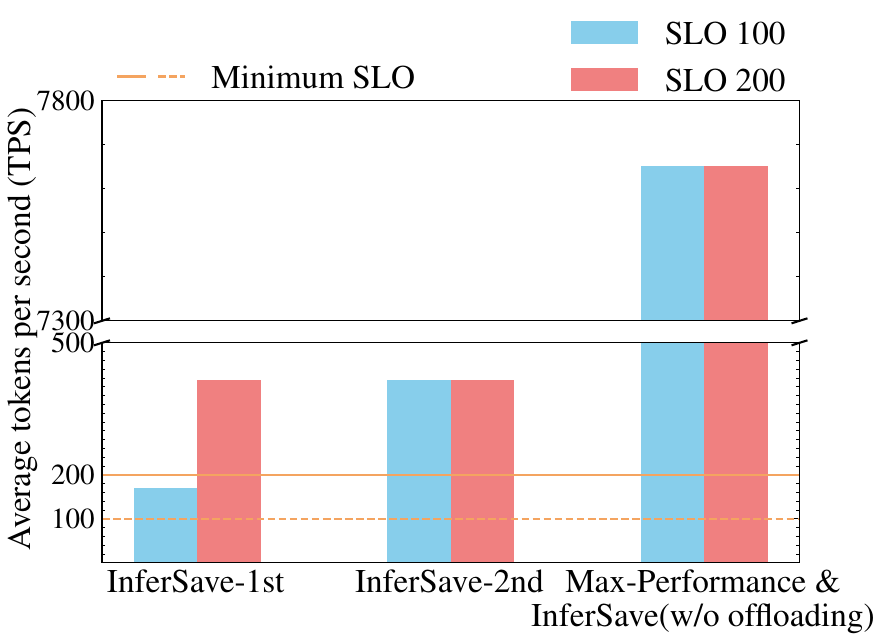}
\end{minipage}
\hfill
\begin{minipage}{0.49\columnwidth} 
    \centering
    \includegraphics[width=\textwidth]{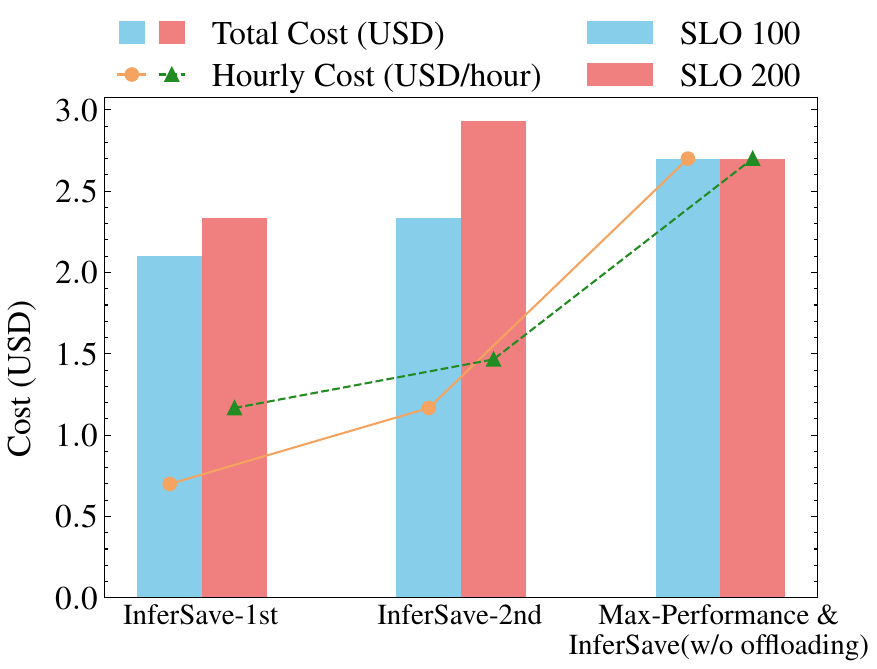}
\end{minipage}
\caption{Comparison of average TPS and cost for different \tbdfixed{} configurations and the baseline configuration under varying SLO constraints for offline inference workloads (Left: Average TPS, Right: Cost).}
\vspace{-8pt}
\label{fig:offline-tps}
\end{figure}

\subsubsection{Offline inference workload results}
Table ~\ref{tab:offline-policy_comparison} and Figure~\ref{fig:offline-tps} shows the instance selection of each policy based on the SLO requirements given for an offline inference workload, and performance and price comparisons from the selection made by each policy. As this workload uses a large input token size, all instances excluding g6e.xlarge make use of KV Cache offloading. Without considering offloading, only one instance can be considered a top choice, and therefore, \tbdfixed{} without offloading chose the same instance as the Max-Performance policy.

Given a SLO requirement of 100 TPS, \tbdfixed{} selected g4dn.xlarge as its top choice, providing a throughput of about 160 TPS with the lowest total processing cost of \$2.13. On the other hand, both Max-Performance and \tbdfixed{} without offloading selected g6e.xlarge, which delivers a very high throughput of about 7600 TPS, but with a total cost of \$2.699, an increase of about 26.7\%. The selection of g6e.xlarge allows for maximum throughput with the ability to store all KV Cache in GPU memory without offloading. However, despite the high throughput and meeting the SLO, the high cost of the instance itself results in lower overall cost efficiency.

With a SLO requirement of 200 TPS, \tbdfixed{} selected g5.xlarge as its top choice, as g4dn.xlarge not longer meet the performance requirements. This instance provides about 400 TPS while maintaining a total cost of \$2.344. On the other hand, the Max-Performance policy still selected g6e.xlarge, providing a performance of about 7600 TPS, but the total cost increased to \$2.699, resulting in about a 15\% higher cost. This shows that without considering offloading, a needlessly highly performant and expensive instance can be chosen, leading to excessively high costs, and thus reducing actual cost efficiency.

\subsubsection{Overall analysis and discussion}
By evaluating experimental results that represent both online chatbot and batch processing workloads, we were able to derive key insights for the efficient operation of LLM inference systems.  
\begin{itemize}
    \item[(i)] \textbf{The impact of a workload's I/O patterns on optimal infrastructure selection:}  
    The requirements of online conversational chatbot inference and batch processing inference differ greatly in input and output token lengths, which act as key factors in determining optimal instance and offloading strategies.
  
    \item[(ii)] \textbf{The significance of selectively applying KV Cache offloading:}
    KV Cache offloading is not a universally applicable strategy for all workloads and achieves the greatest cost reduction when selectively utilized according to workload characteristics. In particular, for offline batch processing workloads with long inputs, cost reductions up to 28\% were possible with KV Cache offloading, while maintaining SLO requirements. On the other hand, in online conversational chatbot workloads, it was often more advantageous to apply KV Cache offloading when considering both cost and performance.
   
    \item[(iii)] \textbf{Finding the optimal interface through \tbdfixed{}:}  
    \tbdfixed{} comprehensively considers the SLO requirements and workload characteristics to find the optimal balance point in cost and performance. Instead of naively selecting the instance with the highest performance, \tbdfixed{} finds the instance with the highest cost efficiency while still satisfying the SLO requirement.
    
\end{itemize}

These results reveal an opportunity for both cost and performance optimization by flexibly adjusting the offloading strategy and GPU instance choice to match workload patterns and SLO requirements. \tbdfixed{} takes this opportunity and performs said optimizations automatically with the selection of the optimal instance by considering information from precise analysis of each workload's characteristics. As a result, \tbdfixed{} achieves optimal performance according to the given SLO while maintaining cost efficiency.

However, this work currently has limitations in that it focuses on a single GPU environment and does not address optimization strategies in multi-GPU or distributed inference environments. We plan to extend \tbdfixed{} to multi-GPU and distributed cluster environments to develop optimization strategies suitable for inference of more complex workloads and large-scale models.

\section{Conclusion}

In this study, we propose \tbdfixed{}, which utilizes SLO-based predictions to automatically select cost-efficient VM instances in the cloud, and validate it across online and offline workloads. We identify opportunities to enhance cost efficiency and utilize cheaper, less powerful GPU instances, while maintaining the specified SLO requirements by exploiting techniques such as KV Cache offloading. Through extensive evaluation across both online and offline inference workloads, our results confirm that \tbdfixed{} accurately exploits said opportunities, and achieves at most 73.7\% lower running costs while maintaining SLO requirements.

This research suggests that LLM service providers can optimize cost and performance in a balanced way by selecting optimal instances based on SLO and effectively utilizing offloading strategies. \tbdfixed{} offers these optimizations in a unified package in a LLM inferencing system that both lowers cost and maintains performance requirements.



{
\setstretch{0.99}
\bibliographystyle{ieeetr}
\bibliography{paper}
}


\end{document}